\documentclass[letterpaper]{article}

\usepackage{aaai18}
\usepackage{times}
\usepackage{helvet}
\usepackage{courier}
\usepackage{url}
\usepackage{graphicx}
\usepackage{amsmath}
\usepackage{amssymb}
\usepackage{siunitx}
\usepackage[utf8]{inputenc}
\usepackage[inline]{enumitem}
\usepackage[nolist,nohyperlinks]{acronym}

\nocopyright

\title{SEE: Towards Semi-Supervised End-to-End Scene Text Recognition}

\author{Christian Bartz \and Haojin Yang \and Christoph Meinel\\
Hasso Plattner Institute, University of Potsdam\\
Prof.-Dr.-Helmert Straße 2-3\\ 14482 Potsdam, Germany\\
\{christian.bartz, haojin.yang, meinel\}@hpi.de
}

\begin{document}
	\maketitle

	\begin{acronym}
		\acro{OCR}{Optical Character Recognition}
		\acro{CNN}{Convolutional Neural Network}
		\acro{RNN}{Recurrent Neural Network}
		\acro{DNN}{Deep Neural Network}
		\acro{BLSTM}{Bidirectional Long-Short Term Memory}
		\acro{LSTM}{Long-Short Term Memory}
		\acro{CTC}{Connectionist Temporal Classification}
		\acro{FSNS}{French Street Name Signs}
		\acro{SGD}{Stochastic Gradient Descent}
	\end{acronym}

	\begin{abstract}
		Detecting and recognizing text in natural scene images is a challenging, yet not completely solved task.
		In recent years several new systems that try to solve at least one of the two sub-tasks (text detection and text recognition) have been proposed.
		In this paper we present SEE, a step towards semi-supervised neural networks for scene text detection and recognition, that can be optimized end-to-end.
		Most existing works consist of multiple deep neural networks and several pre-processing steps.
		In contrast to this, we propose to use a single deep neural network, that learns to detect and recognize text from natural images, in a semi-supervised way.
		SEE is a network that integrates and jointly learns a spatial transformer network, which can learn to detect text regions in an image, and a text recognition network that takes the identified text regions and recognizes their textual content.
		We introduce the idea behind our novel approach and show its feasibility, by performing a range of experiments on standard benchmark datasets, where we achieve competitive results.
	\end{abstract}

	\section{Introduction}
	\label{sec:introduction}

	Text is ubiquitous in our daily lives.
	Text can be found on documents, road signs, billboards, and other objects like cars or telephones.
	Automatically detecting and reading text from natural scene images is an important part of systems, that are to be used for several challenging tasks, such as image-based machine translation, autonomous cars or image/video indexing.
	In recent years the task of detecting text and recognizing text in natural scenes has seen much interest from the computer vision and document analysis community.
	Furthermore, recent breakthroughs \cite{He2016Deep,Jaderberg2015Spatial,Redmon2016You,Ren2015Faster} in other areas of computer vision enabled the creation of even better scene text detection and recognition systems than before \cite{Gomez2017Textproposals,Gupta2016Syntheticb,Shi2016Robust}.
	Although the problem of \ac{OCR} can be seen as solved for text in printed documents, it is still challenging to detect and recognize text in natural scene images.
	Images containing natural scenes exhibit large variations of illumination, perspective distortions, image qualities, text fonts, diverse backgrounds, etc.

	The majority of existing research works developed end-to-end scene text recognition systems that consist of complex two-step pipelines, where the first step is to detect regions of text in an image and the second step is to recognize the textual content of that identified region.
	Most of the existing works only concentrate on one of these two steps.

	\begin{figure}[t]
		\centering
		\includegraphics[width=1.0\linewidth]{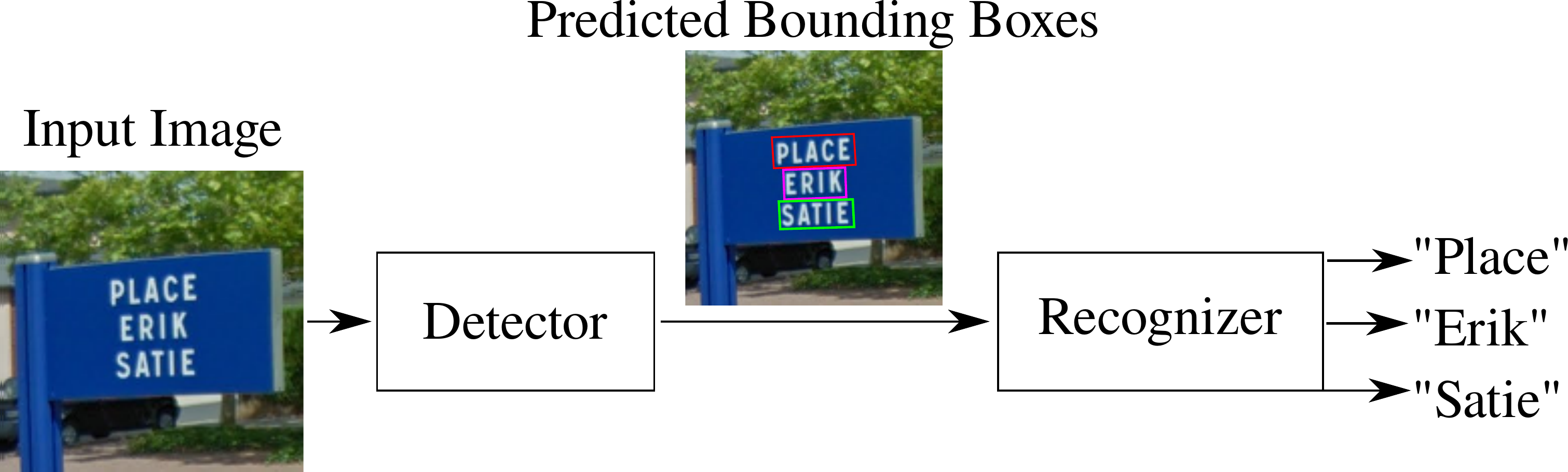}
		\caption{Schematic overview of our proposed system. The input image is fed to a single neural network that consists of a text detection part and a text recognition part. The text detection part learns to detect text in a semi-supervised way, by being jointly trained with the recognition part.}
		\label{fig:schematic_of_system}
	\end{figure}

	In this paper, we present a solution that consists of a single \ac{DNN} that can learn to detect and recognize text in a semi-supervised way.
	In this setting the network only receives the image and the textual labels as input.
	We do not supply any groundtruth bounding boxes.
	The text detection is learned by the network itself.
	This is contrary to existing works, where text detection and text recognition systems are trained separately in a fully-supervised way.
	Recent work \cite{Dai2016InstanceAware} showed that \acp{CNN} are capable of learning how to solve complex multi-task problems, while being trained in an end-to-end manner.
	Our motivation is to use these capabilities of \acp{CNN} and create an end-to-end trainable scene text recognition system, that can be trained on weakly labelled data.
	In order to create such a system, we learn a single \ac{DNN} that is able to find single characters, words or even lines of text in the input image and recognize their content.
	This is achieved by jointly learning a localization network that uses a recurrent spatial transformer \cite{Jaderberg2015Spatial,Snderby2015Recurrent} as attention mechanism and a text recognition network.
	Figure~\ref{fig:schematic_of_system} provides a schematic overview of our proposed system.

	Our contributions are as follows:
	\begin{enumerate*}[label={(\arabic*)}]
		\item We present a novel end-to-end trainable system for scene text detection and recognition by integrating spatial transformer networks.
		\item We propose methods that can improve and ease the work with spatial transformer networks.
		\item We train our proposed system end-to-end, in a semi-supervised way.
		\item We demonstrate that our approach is able to reach competitive performance on standard benchmark datasets.
		\item We provide our code\footnote{\url{https://github.com/Bartzi/see}} and trained models\footnote{\url{https://bartzi.de/research/see}} to the research community.
	\end{enumerate*}

	This paper is structured in the following way:
	We first outline work of other researchers that is related to ours.
	Second, we describe our proposed system in detail.
	We then show and discuss our results on standard benchmark datasets and finally conclude our findings.

	\section{Related Work}
	\label{sec:related_work}

	Over the course of years a rich environment of different approaches to scene text detection and recognition have been developed and published.
	Nearly all systems use a two-step process for performing end-to-end recognition of scene text.
	The first step, is to detect regions of text and extract these regions from the input image.
	The second step, is to recognize the textual content and return the text strings of the extracted text regions.

	It is further possible to divide these approaches into three broad categories:
	\begin{enumerate*}[label={(\arabic*)}]
		\item Systems relying on hand crafted features and human knowledge for text detection and text recognition.
		\item Systems using deep learning approaches, together with hand crafted features, or two different deep networks for each of the two steps.
		\item Systems that do not consist of a two step approach but rather perform text detection and recognition using a single deep neural network.
	\end{enumerate*}
	For each category, we will discuss some of these systems.

	\paragraph{Hand Crafted Features}
		In the beginning, methods based on hand crafted features and human knowledge have been used to perform text detection and recognition.
		These systems used features like MSERs \cite{Neumann2010Method}, Stroke Width Transforms \cite{Epshtein2010Detecting} or HOG-Features \cite{Wang2011EndToEnd} to identify regions of text and provide them to the text recognition stage of the system.
		In the text recognition stage sliding window classifiers \cite{Mishra2012Scene} and ensembles of SVMs \cite{Yao2014Strokelets} or k-Nearest Neighbor classifiers using HOG features \cite{Wang2010Word} were used.
		All of these approaches use hand crafted features that have a large variety of hyper parameters that need expert knowledge to correctly tune them for achieving the best results.

	\paragraph{Deep Learning Approaches}
	More recent systems exchange approaches based on hand crafted features in one or both steps of recognition systems by approaches using \acp{DNN}.
	Gómez and Karatzas \cite{Gomez2017Textproposals} propose a text-specific selective search algorithm that, together with a \ac{DNN}, can be used to detect (distorted) text regions in natural scene images.
	Gupta et al. \cite{Gupta2016Syntheticb} propose a text detection model based on the YOLO-Architecture \cite{Redmon2016You} that uses a fully convolutional deep neural network to identify text regions.

	Bissacco et al. \cite{Bissacco2013Photoocr} propose a complete end-to-end architecture that performs text detection using hand crafted features.
	Jaderberg et al. \cite{Jaderberg2015Reading,Jaderberg2014Deep} propose several systems that use deep neural networks for text detection and text recognition.
	In \cite{Jaderberg2015Reading} Jaderberg et al. propose to use a region proposal network with an extra bounding box regression CNN for text detection.
	A CNN that takes the whole text region as input is used for text recognition.
	The output of this CNN is constrained to a pre-defined dictionary of words, making this approach only applicable to one given language.

	Goodfellow et al. \cite{Goodfellow2014MultiDigit} propose a text recognition system for house numbers, that has been refined by Jaderberg et al. \cite{Jaderberg2014Deep} for unconstrained text recognition.
	This system uses a single \ac{CNN}, taking the whole extracted text region as input, and recognizing the text using one independent classifier for each possible character in the given word.
	Based on this idea He et al. \cite{He2016Reading} and Shi et al. \cite{Shi2016EndToEnd} propose text recognition systems that treat the recognition of characters from the extracted text region as a sequence recognition problem.
	Shi et al. \cite{Shi2016Robust} later improved their approach by firstly adding an extra step that utilizes the rectification capabilities of Spatial Transformer Networks \cite{Jaderberg2015Spatial} for rectifying extracted text lines.
	Secondly they added a soft-attention mechanism to their network that helps to produce the sequence of characters in the input image.
	In their work Shi et al. make use of Spatial Transformers as an extra pre-processing step to make it easier for the recognition network to recognize the text in the image. In our system we use the Spatial Transformer as a core building block for detecting text in a semi-supervised way.

	\paragraph{End-to-End trainable Approaches}
		The presented systems always use a two-step approach for detecting and recognizing text from scene text images.
		Although recent approaches make use of deep neural networks they are still using a huge amount of hand crafted knowledge in either of the steps or at the point where the results of both steps are fused together.
		Smith et al. \cite{Smith2016EndToEnd} and Wojna et al. \cite{Wojna2017AttentionBased} propose an end-to-end trainable system that is able to recognize text on French street name signs, using a single \ac{DNN}.
		In contrast to our system it is not possible for the system to provide the location of the text in the image, only the textual content can be extracted.
		Recently Li et al. \cite{Li2017Towards} proposed an end-to-end system consisting of a single, complex \ac{DNN} that is trained end-to-end and can perform text detection and text recognition in a single forward pass.
		This system is trained using groundtruth bounding boxes and groundtruth labels for each word in the input images, which stands in contrast to our method, where we only use groundtruth labels for each word in the input image, as the detection of text is learned by the network itself.

	\section{Proposed System}
	\label{sec:proposed_system}

	A human trying to find and read text will do so in a sequential manner.
	The first action is to put attention on a word, read each character sequentially and then attend to the next word.
	Most current end-to-end systems for scene text recognition do not behave in that way.
	These systems rather try to solve the problem by extracting all information from the image at once.
	Our system first tries to attend sequentially to different text regions in the image and then recognize their textual content.
	In order to do this, we created a single \ac{DNN} consisting of two stages:
	\begin{enumerate*}[label={(\arabic*)}]
		\item text detection, and
		\item text recognition
	\end{enumerate*}.
	In this section we will introduce the attention concept used by the text detection stage and the overall structure of the proposed system.

	\subsection{Detecting Text with Spatial Transformers}
	\label{subsec:ps_spatial_transformer_networks}

	A spatial transformer proposed by Jaderberg et al. \cite{Jaderberg2015Spatial} is a differentiable module for \acp{DNN} that takes an input feature map $I$ and applies a spatial transformation to this feature map, producing an output feature map $O$.
	Such a spatial transformer module is a combination of three parts.
	The first part is a localization network computing a function $f_{loc}$, that predicts the parameters $\theta$ of the spatial transformation to be applied.
	These predicted parameters are used in the second part to create a sampling grid, which defines a set of points where the input map should be sampled.
	The third part is a differentiable interpolation method, that takes the generated sampling grid and produces the spatially transformed output feature map $O$.
	We will shortly describe each component in the following paragraphs.

	\paragraph{Localization Network}
		The localization network takes the input feature map $I \in \mathbb{R}^{C \times H \times W}$, with $C$ channels, height $H$ and width $W$ and outputs the parameters $\theta$ of the transformation that shall be applied.
		In our system we use the localization network ($f_{loc}$) to predict $N$ two-dimensional affine transformation matrices $A^{n}_{\theta}$, where $n \in \{0, \ldots, N - 1\}$:
		\begin{equation}
			f_{loc}(I) = A^{n}_{\theta} =
			\begin{bmatrix}
				\theta^{n}_1 & \theta^{n}_2 & \theta^{n}_3 \\
				\theta^{n}_4 & \theta^{n}_5 & \theta^{n}_6 \\
			\end{bmatrix}
		\end{equation}

		$N$ is thereby the number of characters, words or textlines the localization network shall localize.
		The affine transformation matrices predicted in that way allow the network to apply translation, rotation, zoom and skew to the input image.

\begin{figure*}[t]
		\centering
		\includegraphics[width=0.9\linewidth]{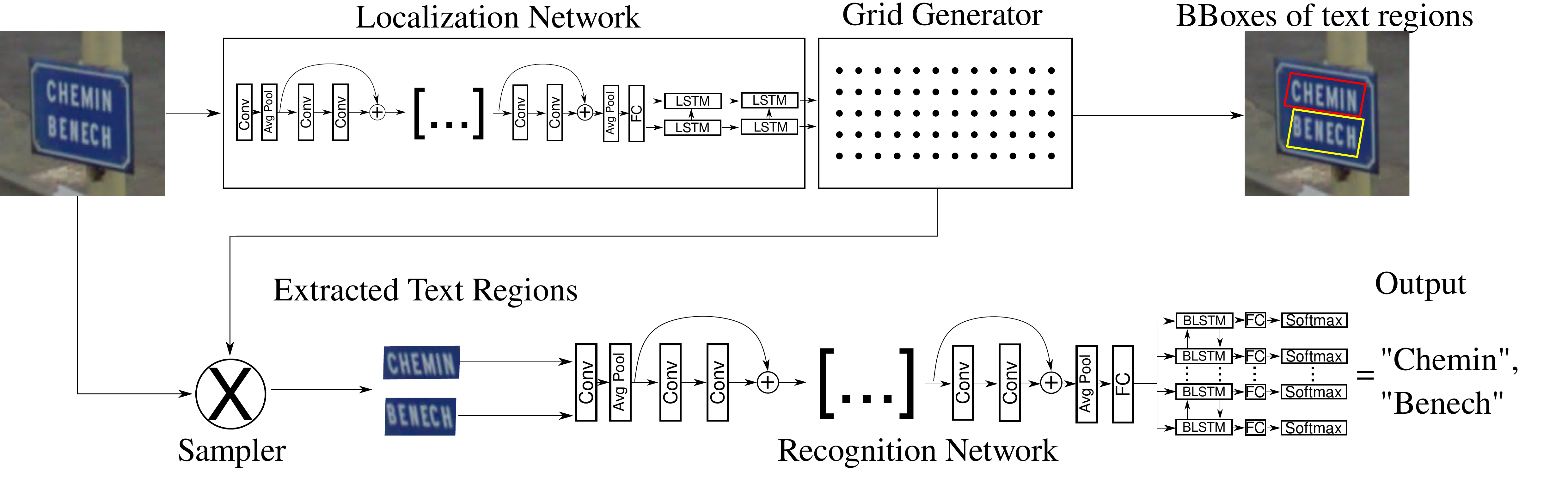}
		\caption{The network used in our work consists of two major parts. The first is the localization network that takes the input image and predicts $N$ transformation matrices, which are used to create $N$ different sampling grids. The generated sampling grids are used in two ways:
		(1) for calculating the bounding boxes of the identified text regions
		(2) for extracting $N$ text regions.
		The recognition network then performs text recognition on these extracted regions.
		The whole system is trained end-to-end by only supplying information about the text labels for each text region.}
		\label{fig:localization_net_structure}
	\end{figure*}

		In our system the $N$ transformation matrices $A^{n}_{\theta}$ are produced by using a feed-forward \ac{CNN} together with a \ac{RNN}. Each of the $N$ transformation matrices is computed using the globally extracted convolutional features $c$ and the hidden state $h_n$ of each time-step of the \ac{RNN}:
		\begin{align}
			c &= f^{conv}_{loc}(I) \\
			h_n &= f^{rnn}_{loc}(c, h_{n-1}) \\
			A^{n}_{\theta} &= g_{loc}(h_n)
		\end{align}
		where $g_{loc}$ is another feed-forward/recurrent network.
		We use a variant of the well known ResNet architecture~\cite{He2016Deep} as \ac{CNN} for our localization network.
		We use this network architecture, because we found that with this network structure our system learns faster and more successfully, as compared to experiments with other network structures, such as the VGGNet \cite{Simonyan2015Very}.
		We argue that this is due to the fact that the residual connections of the ResNet help with retaining a strong gradient down to the very first convolutional layers.
		The \ac{RNN} used in the localization network is a \ac{LSTM} \cite{Hochreiter1997Long} unit.
		This \ac{LSTM} is used to generate the hidden states $h_n$, which in turn are used to predict the affine transformation matrices.
		We used the same structure of the network for all our experiments we report in the next section.
		Figure~\ref{fig:localization_net_structure} provides a structural overview of this network.

\begin{figure}[t]
		\centering
		\includegraphics[width=0.9\linewidth]{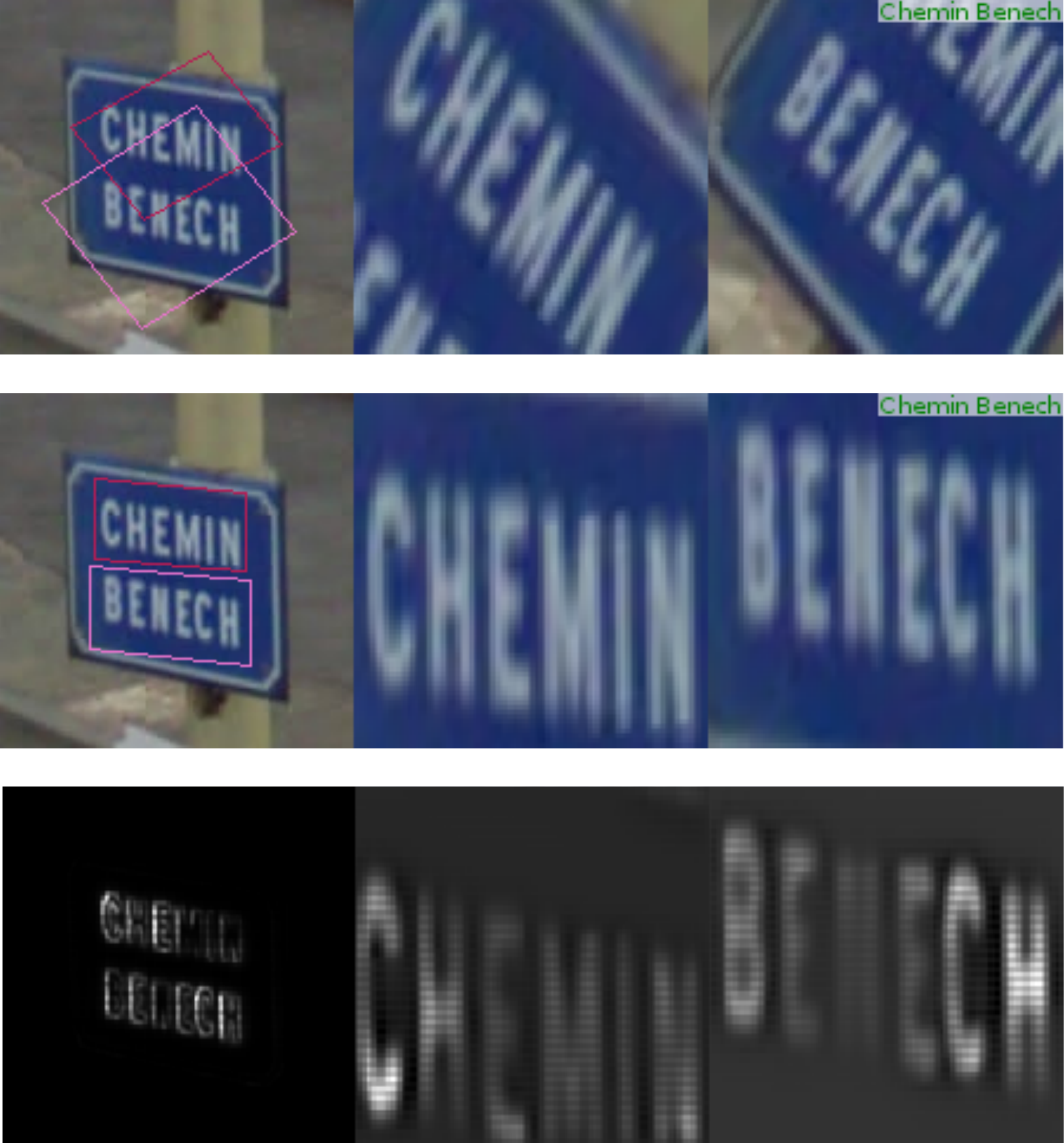}
		\caption{\emph{Top:} predicted bounding boxes of network trained without rotation dropout. \emph{Middle:} predicted bounding boxes of network trained with rotation dropout. \emph{Bottom:} visualization of image parts that have the highest influence on the outcome of the prediction. This visualization has been created using Visualbackprop~\cite{Bojarski2016Visualbackprop}.}
		\label{fig:rotation_dropout_visualization}
	\end{figure}

	\paragraph{Rotation Dropout}
		During our experiments, we found that the network tends to predict transformation parameters, which include excessive rotation.
		In order to mitigate such a behavior, we propose a mechanism that works similarly to dropout~\cite{Srivastava2014Dropout}, which we call rotation dropout.
		Rotation dropout works by randomly dropping the parameters of the affine transformation, which are responsible for rotation.
		This prevents the localization network to output transformation matrices that perform excessive rotation.
		Figure~\ref{fig:rotation_dropout_visualization} shows a comparison of the localization result of a localization network trained without rotation dropout (top) and one trained with rotation dropout (middle).

	\paragraph{Grid Generator}
		The grid generator uses a regularly spaced grid $G_o$ with coordinates $y_{h_o},x_{w_o}$, of height $H_o$ and width $W_o$. The grid $G_o$ is used together with the affine transformation matrices $A^{n}_{\theta}$ to produce $N$ regular grids $G^n$ with coordinates $u^n_{i},v^n_{j}$ of the input feature map $I$, where $i \in H_o$ and $j \in W_o$:
		\begin{equation}
			\begin{pmatrix}
				u^n_{i} \\
				v^n_{j}
			\end{pmatrix}
			= A^{n}_{\theta}
			\begin{pmatrix}
				x_{w_o} \\
				y_{h_o} \\
				1
			\end{pmatrix}
			= \begin{bmatrix}
				\theta^{n}_1 & \theta^{n}_2 & \theta^{n}_3 \\
				\theta^{n}_4 & \theta^{n}_5 & \theta^{n}_6 \\
			\end{bmatrix}
			\begin{pmatrix}
				x_{w_o} \\
				y_{h_o} \\
				1
			\end{pmatrix}
		\end{equation}
		During inference we can extract the $N$ resulting grids $G^n$, which contain the bounding boxes of the text regions found by the localization network.
		Height $H_o$ and width $W_o$ can be chosen freely.

	\paragraph{Localization specific regularizers}
		The datasets used by us, do not contain any samples, where text is mirrored either along the x- or y-axis.
		Therefore, we found it beneficial to add additional regularization terms that penalizes grid, which are mirrored along any axis.
		We furthermore found that the network tends to predict grids that get larger over the time of training, hence we included a further regularizer that penalizes large grids, based on their area.
		Lastly, we also included a regularizer that encourages the network to predict grids that have a greater width than height, as text is normally written in horizontal direction and typically wider than high.
		The main purpose of these localization specific regularizers is to enable faster convergence.
		Without these regularizers, the network will eventually converge, but it will take a very long time and might need several restarts of the training.
		Equation~\ref{eq:loc_loss} shows how these regularizers are used for calculating the overall loss of the network.

	\begin{figure}[t]
		\centering
		\includegraphics[width=0.9\linewidth]{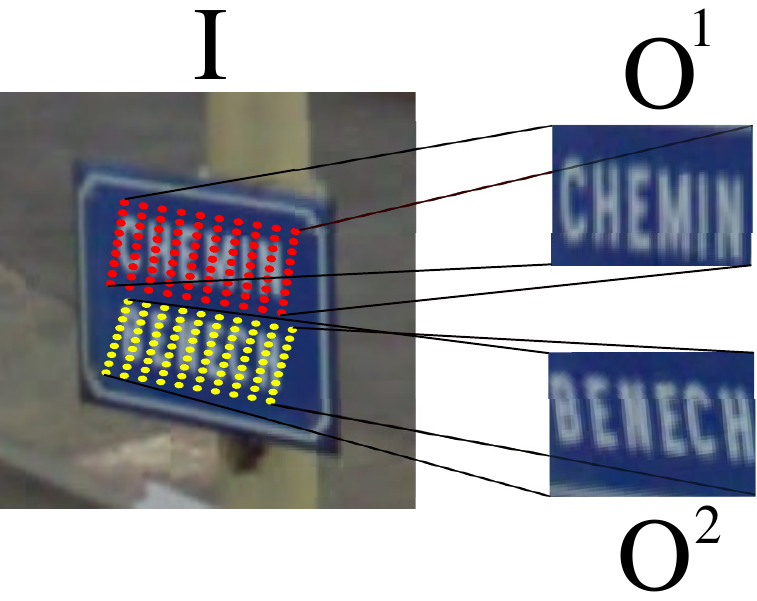}
		\caption{Operation method of grid generator and image sampler. First the grid generator uses the $N$ affine transformation matrices $A^{n}_{\theta}$ to create $N$ equally spaced sampling grids (red and yellow grids on the left side). These sampling grids are used by the image sampler to extract the image pixels at that location, in this case producing the two output images $O^1$ and $O^2$. The corners of the generated sampling grids provide the vertices of the bounding box for each text region, that has been found by the network.}
		\label{fig:transformation_params_overview}
	\end{figure}

	\paragraph{Image Sampling}
		The $N$ sampling grids $G^n$ produced by the grid generator are now used to sample values of the feature map $I$ at the coordinates $u^n_{i},v^n_{j}$ for each $n \in N$. Naturally these points will not always perfectly align with the discrete grid of values in the input feature map.
		Because of that we use bilinear sampling and define the values of the $N$ output feature maps $O^n$ at a given location $i,j$ where $i \in H_o$ and $j \in W_o$ to be:
		\begin{equation}
			O^n_{ij} = \sum^{H}_h \sum^{W}_w I_{hw} max(0, 1 - \lvert u^n_{i} - h \rvert) max(0, 1 - \rvert v^n_{j} - w \rvert)
		\end{equation}
		This bilinear sampling is (sub-)differentiable, hence it is possible to propagate error gradients to the localization network, using standard backpropagation.

	The combination of localization network, grid generator and image sampler forms a spatial transformer and can in general be used in every part of a \ac{DNN}.
	In our system we use the spatial transformer as the first step of our network.
	Figure~\ref{fig:transformation_params_overview} provides a visual explanation of the operation method of grid generator and image sampler.

	\subsection{Text Recognition Stage}

	The image sampler of the text detection stage produces a set of $N$ regions, that are extracted from the original input image.
	The text recognition stage (a structural overview of this stage can be found in Figure~\ref{fig:localization_net_structure}) uses each of these $N$ different regions and processes them independently of each other.
	The processing of the $N$ different regions is handled by a \ac{CNN}.
	This \ac{CNN} is also based on the ResNet architecture as we found that we could only achieve good results, while using a variant of the ResNet architecture for our recognition network.
	We argue that using a ResNet in the recognition stage is even more important than in the detection stage, because the detection stage needs to receive strong gradient information from the recognition stage in order to successfully update the weights of the localization network.
	The \ac{CNN} of the recognition stage predicts a probability distribution $\hat{y}$ over the label space $L_{\epsilon}$, where $L_{\epsilon} = L \cup \{\epsilon\}$, with $L$ being the alphabet used for recognition, and $\epsilon$ representing the blank label.
	The network is trained by running a \ac{LSTM} for a fixed number of $T$ timesteps and calculating the cross-entropy loss for the output of each timestep.
	The choice of number of timesteps $T$ is based on the number of characters, of the longest word, in the dataset.
	The loss $\mathcal{L}$ is computed as follows:
	\begin{align}
		\mathcal{L}_{grid}^n &= \lambda_1 \times \mathcal{L}_{ar}(G^n) + \lambda_2 \times \mathcal{L}_{as}(G^n) + \mathcal{L}_{di}(G^n) \label{eq:loc_loss} \\
		\mathcal{L} &= \sum^{N}_{n=1} (\sum^{T}_{t = 1} (P(l_t^n | O^n)) + \mathcal{L}_{grid}^n)
	\end{align}
	Where $\mathcal{L}_{ar}(G^n)$ is the regularization term based on the area of the predicted grid $n$, $\mathcal{L}_{as}(G^n)$ is the regularization term based on the aspect ratio of the predicted grid $n$, and $\mathcal{L}_{di}(G^n)$ is the regularization term based on the direction of the grid $n$, that penalizes mirrored grids.
	$\lambda_1$ and $\lambda_2$ are scaling parameters that can be chosen freely.
	The typical range of these parameters is $0 < \lambda_1,\lambda_2 < 0.5$.
	$l_t^n$ is the label $l$ at time step $t$ for the n-th word in the image.

	\subsection{Model Training}

	The training set $X$ used for training the model consists of a set of input images $I$ and a set of text labels $L_I$ for each input image.
	We do not use any labels for training the text detection stage.
	The text detection stage is learning to detect regions of text by using only the error gradients, obtained by calculating the cross-entropy loss, of the predictions and the textual labels, for each character of each word.
	During our experiments we found that, when trained from scratch, a network that shall detect and recognize more than two text lines does not converge.
	In order to overcome this problem we designed a curriculum learning strategy~\cite{Bengio2009Curriculum} for training the system.
	The complexity of the supplied training images under this curriculum is gradually increasing, once the accuracy on the validation set has settled.

	During our experiments we observed that the performance of the localization network stagnates, as the accuracy of the recognition network increases.
	We found that restarting the training with the localization network initialized using the weights obtained by the last training and the recognition network initialized with random weights, enables the localization network to improve its predictions and thus improve the overall performance of the trained network.
	We argue that this happens because the values of the gradients propagated to the localization network decrease, as the loss decreases, leading to vanishing gradients in the localization network and hence nearly no improvement of the localization.

	\section{Experiments}
	\label{sec:experiments}

	In this section we evaluate our presented network architecture on standard scene text detection/recognition benchmark datasets.
	While performing our experiments we tried to answer the following questions:
	\begin{enumerate*}[label={(\arabic*)}]
		\item Is the concept of letting the network automatically learn to detect text feasible?
		\item Can we apply the method on a real world dataset?
		\item Can we get any insights on what kind of features the network is trying to extract?
	\end{enumerate*}

	In order to answer these questions, we used different datasets.
	On the one hand we used standard benchmark datasets for scene text recognition.
	On the other hand we generated some datasets on our own.
	First, we performed experiments on the SVHN dataset \cite{Netzer2011Reading}, that we used to prove that our concept as such is feasible.
	Second, we generated more complex datasets based on SVHN images, to see how our system performs on images that contain several words in different locations.
	The third dataset we exerimented with, was the \acf{FSNS} dataset \cite{Smith2016EndToEnd}.
	This dataset is the most challenging we used, as it contains a vast amount of irregular, low resolution text lines, that are more difficult to locate and recognize than text lines from the SVHN datasets.
	We begin this section by introducing our experimental setup.
	We will then present the results and characteristics of the experiments for each of the aforementioned datasets.
	We will conclude this section with a brief explanation of what kinds of features the network seems to learn.

	\subsection{Experimental Setup}
	\label{ssec:experimental_setup}

	\paragraph{Localization Network}
	The localization network used in every experiment is based on the ResNet architecture \cite{He2016Deep}.
	The input to the network is the image where text shall be localized and later recognized.
	Before the first residual block the network performs a $3 \times 3$ convolution, followed by batch normalization~\cite{Ioffe2015Batcha}, ReLU~\cite{Nair2010Rectified}, and a $2 \times 2$ average pooling layer with stride 2.
	After these layers three residual blocks with two $3 \times 3$ convolutions, each followed by batch normalization and ReLU, are used.
	The number of convolutional filters is 32, 48 and 48 respectively.
	A $2 \times 2$ max-pooling with stride 2 follows after the second residual block.
	The last residual block is followed by a $5 \times 5$ average pooling layer and this layer is followed by a \ac{LSTM} with 256 hidden units.
	Each time step of the \ac{LSTM} is fed into another \ac{LSTM} with 6 hidden units.
	This layer predicts the affine transformation matrix, which is used to generate the sampling grid for the bilinear interpolation.
	We apply rotation dropout to each predicted affine transformation matrix, in order to overcome problems with excessive rotation predicted by the network.

	\paragraph{Recognition Network}
	The inputs to the recognition network are $N$ crops from the original input image, representing the text regions found by the localization network.
	In our SVHN experiments, the recognition network has the same structure as the localization network, but the number of convolutional filters is higher.
	The number of convolutional filters is 32, 64 and 128 respectively.
	We use an ensemble of $T$ independent softmax classifiers as used in \cite{Goodfellow2014MultiDigit} and \cite{Jaderberg2014Deep} for generating our predictions.
	In our experiments on the FSNS dataset we found that using ResNet-18~\cite{He2016Deep} significantly improves the obtained recognition accuracies.

	\paragraph{Alignment of Groundtruth}
	During training we assume that all groundtruth labels are sorted in western reading direction, that means they appear in the following order:
	\begin{enumerate*}[label={\arabic*.}]
		\item from top to bottom, and
		\item from left to right.
	\end{enumerate*}
	We stress that currently it is very important to have a consistent ordering of the groundtruth labels, because if the labels are in a random order, the network rather predicts large bounding boxes that span over all areas of text in the image.
	We hope to overcome this limitation, in the future, by developing a method that allows random ordering of groundtruth labels.

	\paragraph{Implementation}
	We implemented all our experiments using Chainer \cite{chainer_learningsys2015}. We conducted all our experiments on a work station which has an Intel(R) Core(TM) i7-6900K CPU, 64 GB RAM and 4 TITAN X (Pascal) GPUs.

	\subsection{Experiments on the SVHN dataset}
	\label{ssec:svhn_experiments}

	With our first experiments on the SVHN dataset \cite{Netzer2011Reading} we wanted to prove that our concept works.
	We therefore first conducted experiments, similar to the experiments in \cite{Jaderberg2015Spatial}, on SVHN image crops with a single house number in each image crop, that is centered around the number and also contains background noise.
	Table~\ref{tab:svhn_results} shows that we are able to reach competitive recognition accuracies.

	\begin{table}
		\begin{center}
			\begin{tabular}{|l|c|}
				\hline
				Method & 64px \\
				\hline
				\cite{Goodfellow2014MultiDigit} & \SI{96.0}{\percent} \\
				\cite{Jaderberg2015Spatial} & \SI{96.3}{\percent} \\
				\hline
				Ours & \SI{95.2}{\percent} \\
				\hline
			\end{tabular}
		\end{center}
		\caption{Sequence recognition accuracies on the SVHN dataset. When recognizing house number on crops of $64 \times 64$ pixels, following the experimental setup of \cite{Goodfellow2014MultiDigit}}
		\label{tab:svhn_results}
	\end{table}

\begin{figure}[t]
		\centering
		\includegraphics[width=0.9\linewidth]{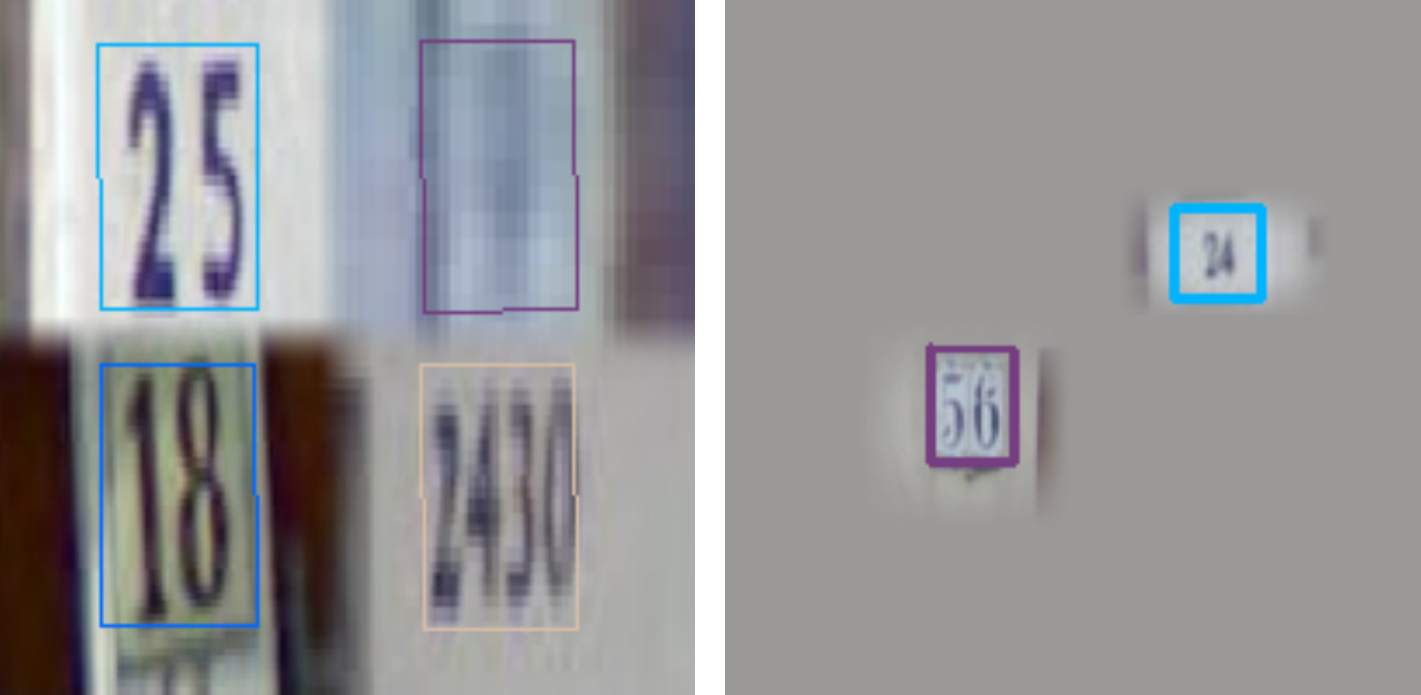}
		\caption{Samples from our generated datasets, including bounding boxes predicted by our model. \textit{Left:} Sample from regular grid dataset, \textit{Right:} Sample from dataset with randomly positioned house numbers.}
		\label{fig:svhn_grid_dataset}
	\end{figure}

	Based on this experiment we wanted to determine whether our model is able to detect different lines of text that are arranged in a regular grid, or placed at random locations in the image.
	In Figure~\ref{fig:svhn_grid_dataset} we show samples from our two generated datasets, that we used for our other experiments based on SVHN data.
	We found that our network performs well on the task of finding and recognizing house numbers that are arranged in a regular grid.

	During our experiments on the second dataset, created by us, we found that it is not possible to train a model from scratch, which can find and recognize more than two textlines that are scattered across the whole image.
	We therefore resorted to designing a curriculum learning strategy that starts with easier samples first and then gradually increases the complexity of the train images.

	\begin{figure*}[ht!]
		\centering
		\includegraphics[width=0.99\linewidth]{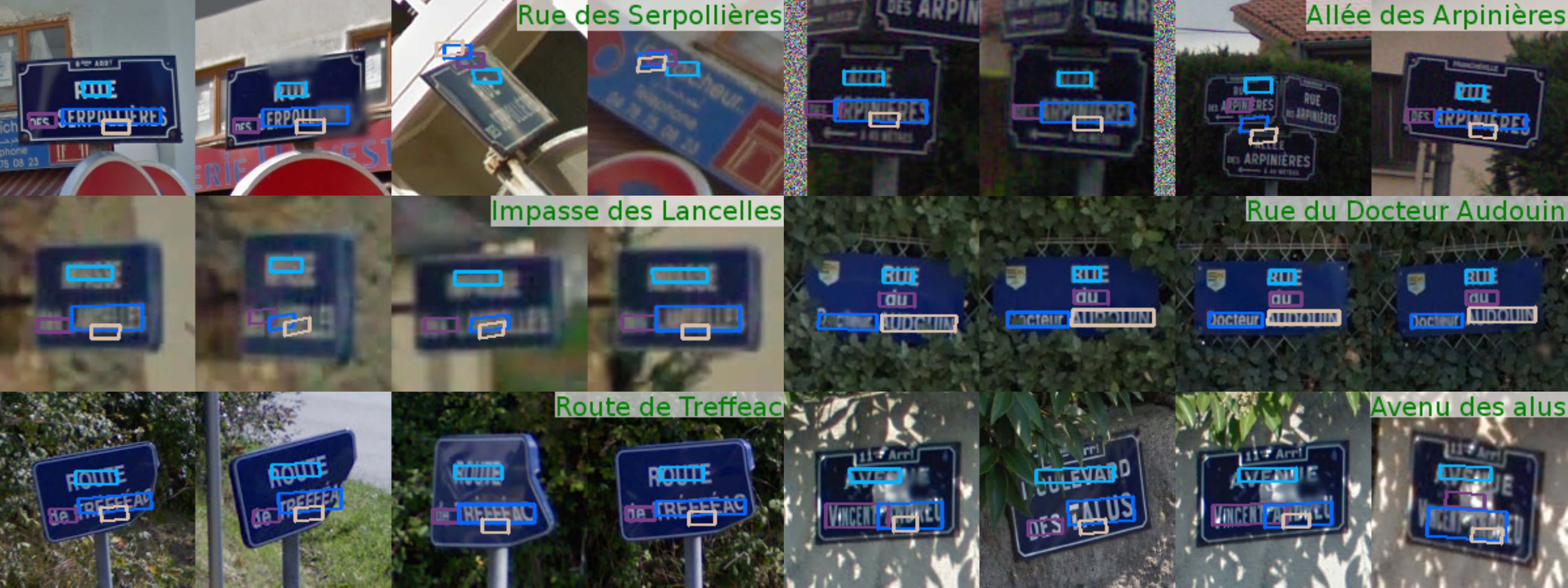}
		\caption{Samples from the \ac{FSNS} dataset, these examples show the variety of different samples in the dataset and also how well our system copes with these samples. The bottom row shows two samples, where our system fails to recognize the correct text. The right image is especially interesting, as the system here tries to mix information, extracted from two different street signs, that should not be together in one sample.}
		\label{fig:fsns_examples}
	\end{figure*}

	\subsection{Experiments on the \ac{FSNS} dataset}
	\label{ssec:fsns_experiments}

	Following our scheme of increasing the difficulty of the task that should be solved by the network, we chose the \acf{FSNS} dataset by Smith et al. \cite{Smith2016EndToEnd} to be our next dataset to perform experiments on.
	The \ac{FSNS} dataset contains more than 1 million images of French street name signs, which have been extracted from Google Streetview.
	This dataset is the most challenging dataset for our approach as it
	\begin{enumerate*}[label={(\arabic*)}]
		\item contains multiple lines of text with varying length, which are embedded in natural scenes with distracting backgrounds, and
		\item contains a lot of images where the text is occluded, not correct, or nearly unreadable for humans.
	\end{enumerate*}

	During our first experiments with that dataset, we found that our model is not able to converge, when trained on the supplied groundtruth.
	We argue that this is because our network was not able to learn the alignment of the supplied labels with the text in the images of the dataset.
	We therefore chose a different approach, and started with experiments where we tried to find individual words instead of textlines with more than one word.
	Table~\ref{tab:fsns_results} shows the performance of our proposed system on the \ac{FSNS} benchmark dataset. We are currently able to achieve competitive performance on this dataset. We are still behind the results reported by Wojna et al.~\cite{Wojna2017AttentionBased}. This likely due to the fact that we used a feature extractor that is weaker (ResNet-18) compared to the one used by Wojna et al. (Inception-ResNet v2). Also recall that our method is not only able to determine the text in the images, but also able to extract the location of the text, although we never explicitly told the network where to find the text! The network learned this completely on its own in a semi-supervised manner.

	\begin{table}
		\centering
		\begin{tabular}{|l|c|}
			\hline
			Method & Sequence Accuracy \\
			\hline
			\cite{Smith2016EndToEnd} & \SI{72.5}{\percent} \\
			\cite{Wojna2017AttentionBased} & \SI{84.2}{\percent} \\
			\hline
			Ours & \SI{78.0}{\percent} \\
			\hline
		\end{tabular}
		\caption{Recognition accuracies on the \ac{FSNS} benchmark dataset.}
		\label{tab:fsns_results}
	\end{table}

	\subsection{Insights}

	During the training of our networks, we used Visualbackprop~\cite{Bojarski2016Visualbackprop} to visualize the regions that the network deems to be the most interesting.
	Using this visualization technique, we could observe that our system seems to learn different types of features for each subtask.
	Figure~\ref{fig:rotation_dropout_visualization} (bottom) shows that the localization network learns to extract features that resemble edges of text and the recognition network learns to find strokes of the individual characters in each cropped word region.
	This is an interesting observation, as it shows that our \ac{DNN} tries to learn features that are closely related to the features used by systems based on hand-crafted features.

	\section{Conclusion}
	\label{sec:conclusion}

	In this paper we presented a system that can be seen as a step towards solving end-to-end scene text recognition, only using a single multi-task deep neural network.
	We trained the text detection component of our model in a semi-supervised way and are able to extract the localization results of the text detection component.
	The network architecture of our system is simple, but it is not easy to train this system, as a successful training requires a clever curriculum learning strategy.
	We also showed that our network architecture can be used to reach competitive results on different public benchmark datasets for scene text detection/recognition.

	At the current state we note that our models are not fully capable of detecting text in arbitrary locations in the image, as we saw during our experiments with the \ac{FSNS} dataset.
	Right now our model is also constrained to a fixed number of maximum words that can be detected with one forward pass. In our future work, we want to redesign the network in a way that makes it possible for the network to determine the number of textlines in an image by itself.

	\fontsize{9.5pt}{10.5pt}
	\selectfont
	\bibliography{Remote}
	\bibliographystyle{aaai}

\end{document}